\documentclass[letterpaper, 10 pt, conference]{ieeeconf}  

\IEEEoverridecommandlockouts                              
\usepackage{graphicx}
\usepackage{multirow}
\usepackage{blindtext}
\usepackage{subcaption}
\usepackage{xcolor}
\usepackage{adjustbox}
\usepackage[normalem]{ulem}
\usepackage{float}
\usepackage{cite}
\title{\LARGE \bf
3D Reconstruction of Whole Stomach from Endoscope Video\\Using Structure-from-Motion
}

\author{Aji Resindra Widya$^{1}$, Yusuke Monno$^{1}$, Kosuke Imahori$^{1}$, Masatoshi Okutomi$^{1}$,\\ Sho Suzuki$^{2}$, Takuji Gotoda$^{2}$, and Kenji Miki$^{3}$ 
\thanks{This work was partly supported by JSPS KAKENHI Grant Number 17H00744.}
\thanks{$^{1}$A. R. Widya, Y. Monno, K. Imahori, and M. Okutomi are with the Department of Systems and Control Engineering, School of Engineering, Tokyo Institute of Technology, Meguro-ku, Tokyo 152-8550, Japan
(e-mail: aresindra@ok.sc.e.titech.ac.jp; ymonno@ok.sc.e.titech.ac.jp; kimahori@ok.sc.e.titech.ac.jp; mxo@sc.e.titech.ac.jp).}
\thanks{$^{2}$S. Suzuki and T. Gotoda are with the Division of Gastroenterology and Hepatology, Department of Medicine, Nihon University School of Medicine, Chiyoda-ku, Tokyo 101-8309, Japan.}
\thanks{$^{3}$K. Miki is with the Department of Internal Medicine, Tsujinaka Hospital Kashiwanoha, Kashiwa-city, Chiba 277-0871, Japan.}%
}

\begin{document}

\maketitle
\thispagestyle{empty}
\pagestyle{empty}

\begin{abstract}
Gastric endoscopy is a common clinical practice that enables medical doctors to diagnose the stomach inside a body. 
In order to identify a gastric lesion's location such as early gastric cancer within the stomach, this work addressed to reconstruct the 3D shape of a whole stomach with color texture information generated from a standard monocular endoscope video. Previous works have tried to reconstruct the 3D structures of various organs from endoscope images. However, they are mainly focused on a partial surface. In this work, we investigated how to enable structure-from-motion~(SfM) to reconstruct the whole shape of a stomach from a standard endoscope video. We specifically investigated the combined effect of chromo-endoscopy and color channel selection on SfM. Our study found that 3D reconstruction of the whole stomach can be achieved by using red channel images captured under chromo-endoscopy by spreading indigo carmine (IC) dye on the stomach surface.
\end{abstract}

\section{Introduction}

Gastric endoscopy is a well-adopted procedure that enables medical doctors to diagnose the stomach inside a body. However, there still exists some challenges to doctors such as the limited point of view and the uncertainty of endoscope poses relative to a target organ.
The accurate localization of a malignant lesion within the global view of the whole stomach is crucial for gastric surgeons to decide the operative procedure of the laparoscopic gastrectomy for early gastric cancer. The location of the malignant lesion is usually identified by the double contrast barium radiography~\cite{yamamichi2016comparative}. However, morphological evaluations such as barium study sometimes cause the gastric surgeons difficulty in identifying flat malignant lesions. Recently, 3D computed tomography (CT) gastrography was developed for the lesion localization purpose~\cite{kim2015role}. However, 3D CT gastrography does not embed color texture information to the reconstructed 3D model. If the 3D shape of the whole stomach can be reconstructed from a standard endoscopic video, the location of the malignant lesion can be identified by the visual color information in addition to the 3D morphological information, which should be very valuable for the gastric surgeons.

Previous studies have shown that 3D endoscopy systems (e.g., a stereo endoscope) have advantages over traditional 2D endoscopes in fields such as computer-aided laparoscopic surgery~\cite{maier_optical_2013} and endoscopic surface imaging~\cite{geng2014review}. Nevertheless, those 3D systems are not widely available and the 2D counterpart is still the mainstream.

Some existing works have proposed a software solution to reconstruct the 3D structure of a target organ (e.g., colon, liver, and larynx) with the estimated endoscope poses from an endoscope video. The methods are ranging from shape-from-shading~(SfS)~\cite{okatani1997shape,henrique2000towards}, visual simultaneous localization and mapping~(SLAM)~\cite{grasa2014visual,mahmoud2016orbslam,Mahmoud2019live}, and structure-from-motion~(SfM)~\cite{mills2014hierarchical,sun2013surface,lurie20173d,furukawa2016shape,schmalz2012endoscopic,alcantarilla2013enhanced}. Even though SfS can reconstruct an organ's surface from a single image, it requires accurate estimation of the light position, which is a difficult problem. SLAM offers a real-time solution with the reconstruction quality as a trade-off. SLAM uses a simple feature detector and descriptor and also sequential feature matching, which leads to a limited reconstruction quality. On the other hand, SfM offers an off-line solution with higher reconstruction quality. SfM uses a more accurate feature detector and descriptor to obtain higher quality features. Moreover, SfM can exhaustively use all input images to find feature correspondences and perform global reconstruction optimization applying bundle adjustment. However, since SfM relies on the detected features, it is still challenging to reconstruct texture-less surfaces, which are common in internal organs. To tackle this challenge, some systems~\cite{furukawa2016shape,schmalz2012endoscopic} exploit a projector to add a structured light pattern on the texture-less surface. Although these systems can successfully increase the number of features, they requires expensive hardware modification. Enhanced imaging colonoscopy and narrow-band imaging were also applied to enhance the surface details for SfM~\cite{alcantarilla2013enhanced}. The above-mentioned works only demonstrated the reconstruction results of a partial surface, which is not sufficient for many potential applications such as the 3D localization of a lesion within the whole shape of the organ.

In this work, we aimed at reconstructing the 3D model of a whole stomach with color texture information from a standard endoscopic video using SfM. We specifically investigated the combined effect of chromo-endoscopy and color channel selection on SfM to achieve better reconstruction quality. Our study found that 3D reconstruction of the whole stomach can be achieved by using red channel images captured under chromo-endoscopy by spreading indigo carmine (IC) dye onto the stomach surface. To the best of our knowledge, this is the first paper to report a successful 3D reconstruction of a whole stomach and visualize the color details of the mucosal surface of it by texture mapping generated from the standard monocular endoscope video.

We also demonstrate our custom viewer that can visualize a particular image frame's location in the 3D model.

\section{Materials and Methods}

In this section, we briefly describe the data collection and the 3D reconstruction method. We first explain our endoscopy hardware setup and the captured video sequences information (Section~\ref{sec:datacollection}). Then, we explain each component of our method starting from the input images extraction for SfM (Section~\ref{sec:dataproc}), the SfM pipeline (Section~\ref{sec:reconstruct}), and the mesh and texture generation (Section~\ref{sec:meshrecon}).

\subsection{Data collection} \label{sec:datacollection}
This study was conducted in accordance with the Declaration of Helsinki. The Institutional Review Board at Nihon University Hospital approved the study protocol on March 8, 2018, before patient recruitment. Informed consent was obtained from all patients before they were enrolled. This study was registered with the University Hospital Medical Information Network (UMIN) Clinical Trials Registry (identification No.: UMIN000031776) on March 17, 2018. This study was also approved by the research ethics committee of Tokyo Institute of Technology, where 3D reconstruction experiments were conducted.

We captured the endoscope video using a standard endoscope system. We used an Olympus IMH-20 image management hub coupled with a GIF-H290 scope. To prevent any compression and unwanted artifacts such as image interlacing, we used an Ephipan video grabber to capture unprocessed data from the image management hub. The video data was saved as an AVI format in 30 frames per second with full HD resolution.

The videos used for 3D reconstruction were captured on three different subjects undergoing general gastrointestinal endoscopy. As shown in Figure~\ref{fig:colorartifact} (a) and (b), each video contains two image sequences captured without and with spraying the IC blue color dye onto the stomach surface as chromo-endoscopy, which is widely applied in endoscopy to enhance the surface visualization. For the dye, we used $\textrm{C}_{16}\textrm{H}_{8}\textrm{N}_{2}\textrm{Na}_{2}\textrm{O}_{8}\textrm{S}_{2}$ manufactured by Daiichi Sankyo Company, Limited, Tokyo, Japan. Additionally, we captured images of a planar checkerboard pattern from multiple orientations for the camera calibration purpose. 

\subsection{Pre-processing of the collected data}\label{sec:dataproc}
The pre-processing of the collected data is performed to estimate intrinsic camera parameters and to extract input images for SfM. This process includes camera calibration, frame extraction, and color channel separation as follows.

An endoscopy camera generally uses an ultra-wide lens to provide a large angle of view inside the stomach. As a trade-off, the ultra wide lens introduces a strong visual distortion and produces images with a convex non-rectilinear appearance, which leads to incorrectly estimated 3D structure. Therefore, camera calibration is needed to obtain the intrinsic camera parameters such as focal length, projection center, and distortion parameters. We used the previously captured planar checkerboard pattern images and a fish-eye camera model~\cite{kannala2006generic} for the camera calibration. The acquired intrinsic camera parameters were used to optimize the 3D reconstruction process in SfM and to correct the image's distortion.

\begin{figure}[t!]
\centering
    \vspace{1mm}
    \begin{subfigure}{0.49\columnwidth}
    \centering
    \includegraphics[width=\columnwidth]{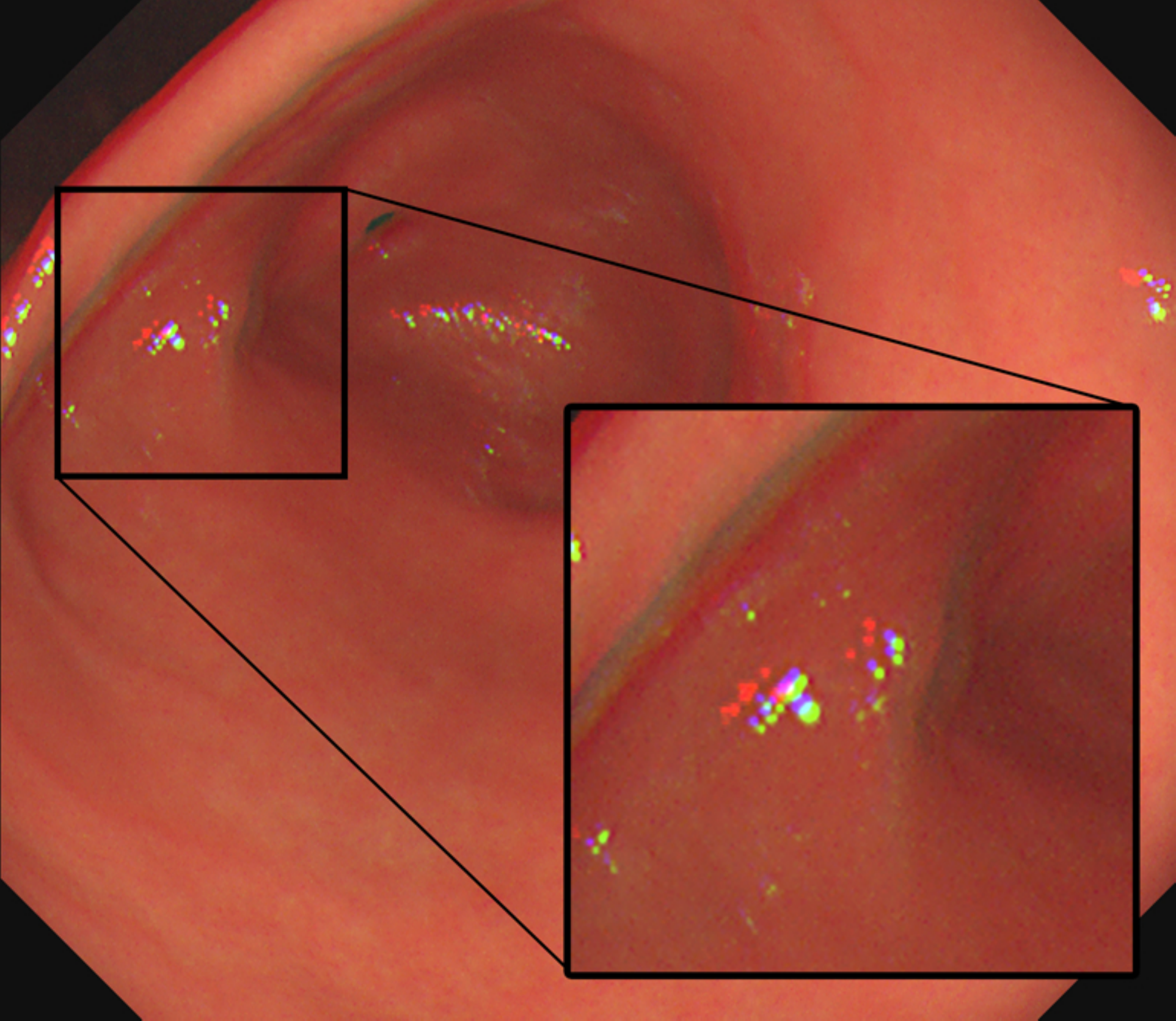}\\ \vspace{0.5mm}
    \caption{RGB without IC dye}
    \label{subfig:RGBCE}
    \end{subfigure}\hspace{\fill}
    \begin{subfigure}{0.49\columnwidth}
    \centering
    \includegraphics[width=\columnwidth]{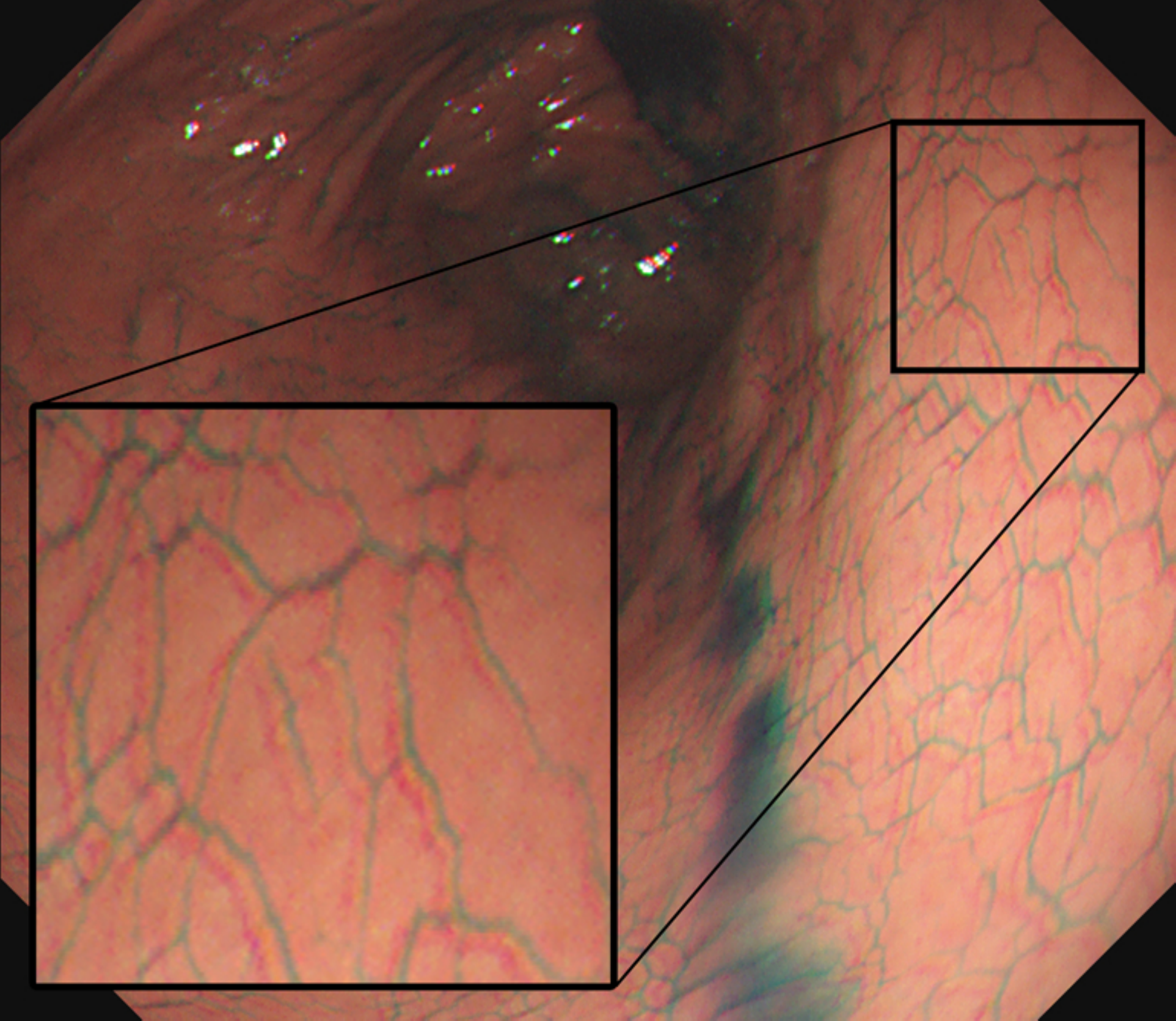}\\ \vspace{0.5mm}
    \caption{RGB with IC dye}
    \end{subfigure}
    
    \begin{subfigure}{0.3\columnwidth}
    \centering
    \vspace{1mm}
    \includegraphics[width=\columnwidth]{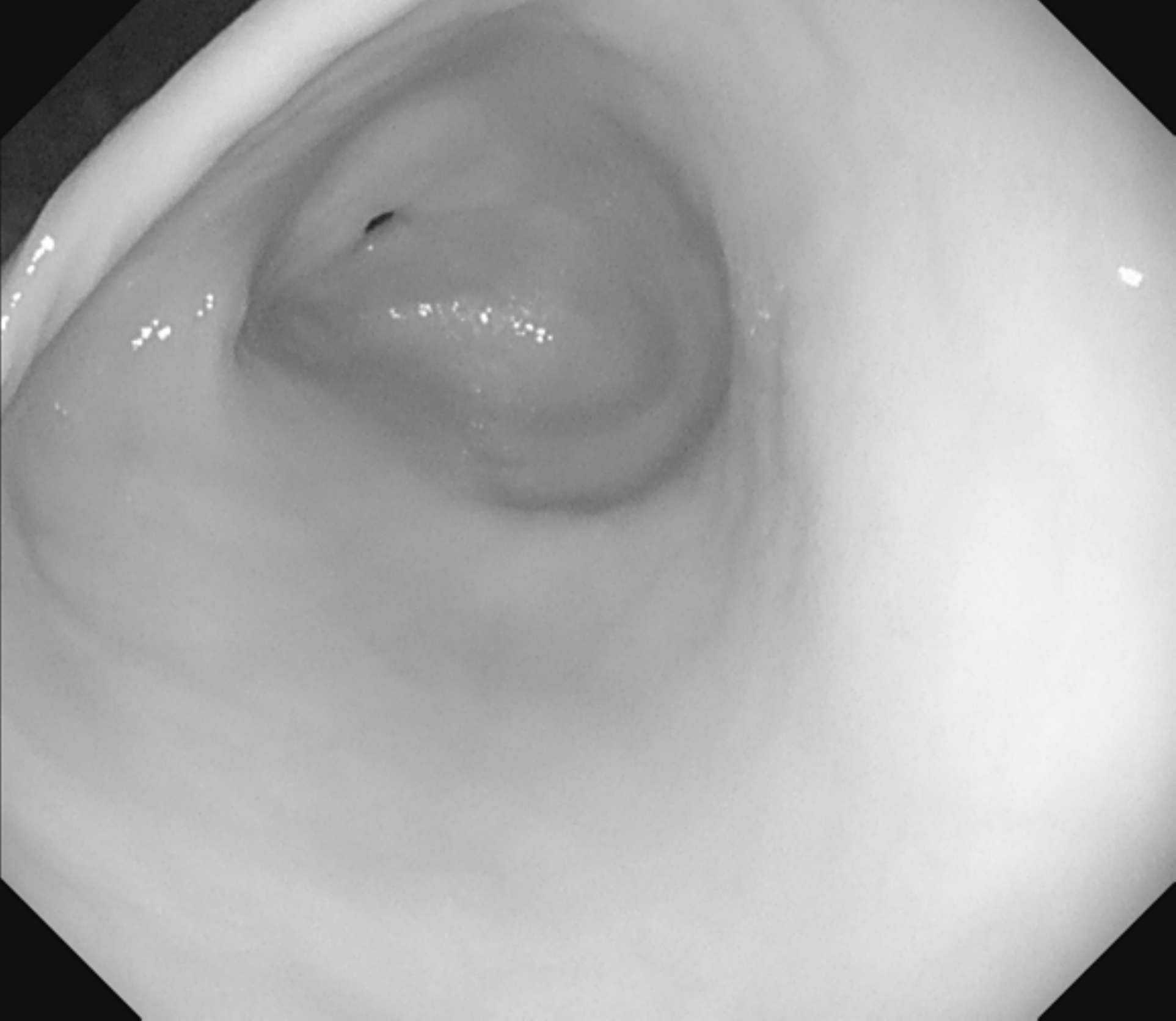}\\ \vspace{0.5mm}
    \caption{Red of (a)}
    \end{subfigure}\hspace{\fill}
    \begin{subfigure}{0.3\columnwidth}
    \centering
    \vspace{1mm}
    \includegraphics[width=\columnwidth]{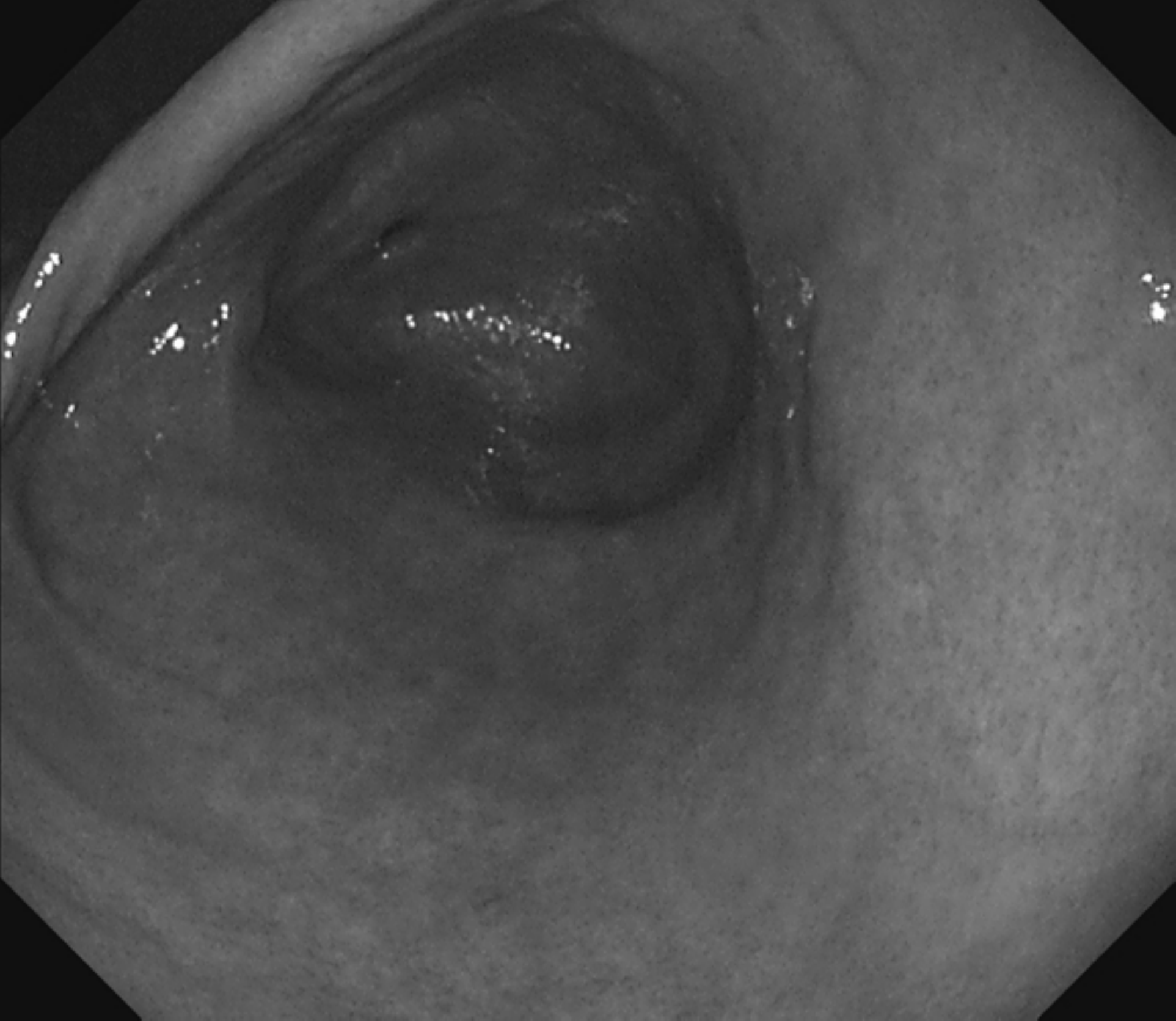}\\ \vspace{0.5mm}
    \caption{Green of (a)}
    \end{subfigure}\hspace{\fill}
    \begin{subfigure}{0.3\columnwidth}
    \centering
    \vspace{1mm}
    \includegraphics[width=\columnwidth]{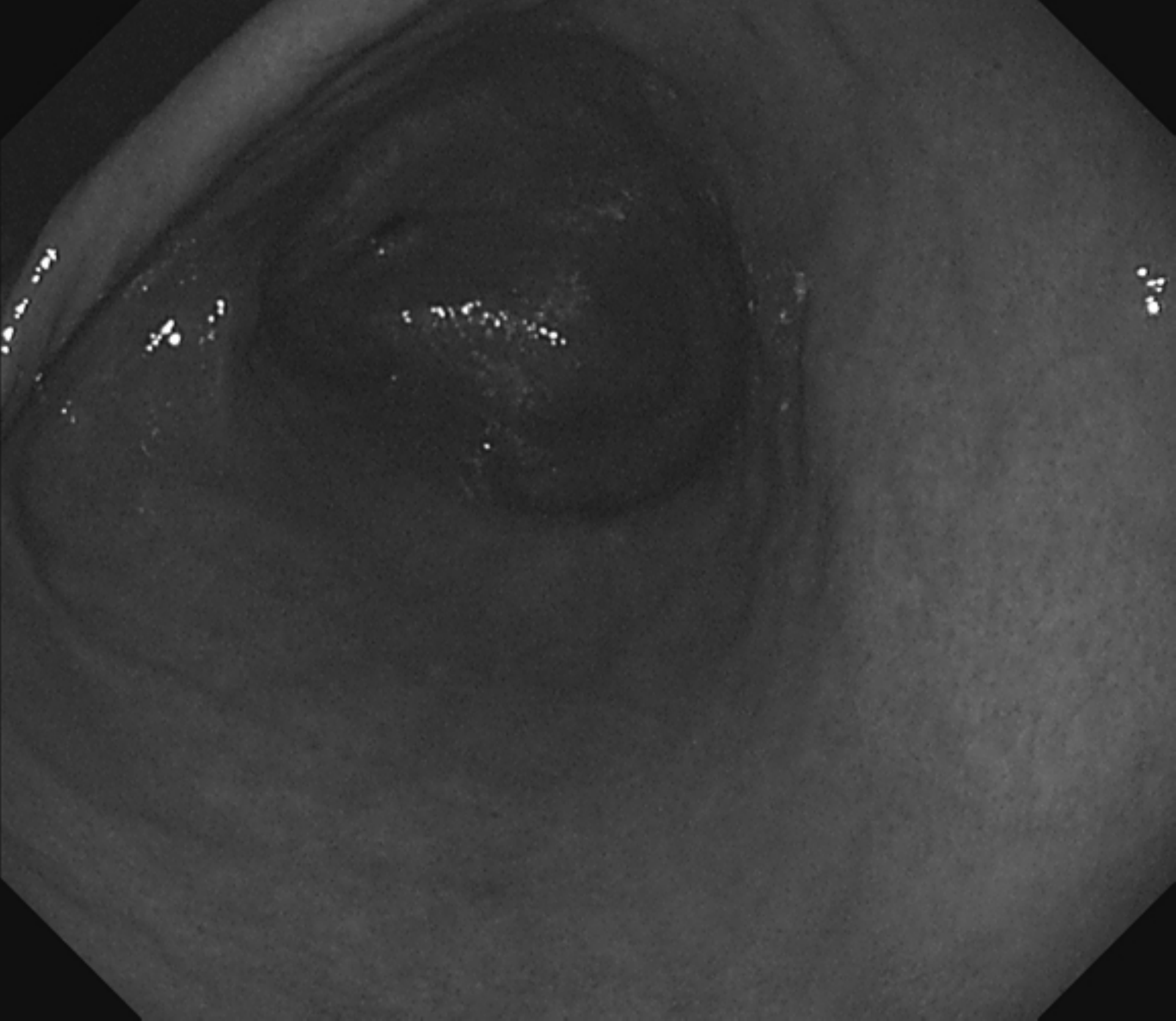}\\ \vspace{0.5mm}
    \caption{Blue of (a)}
    \end{subfigure}
    
    \begin{subfigure}{0.3\columnwidth}
    \centering
    \vspace{1mm}
    \includegraphics[width=\columnwidth]{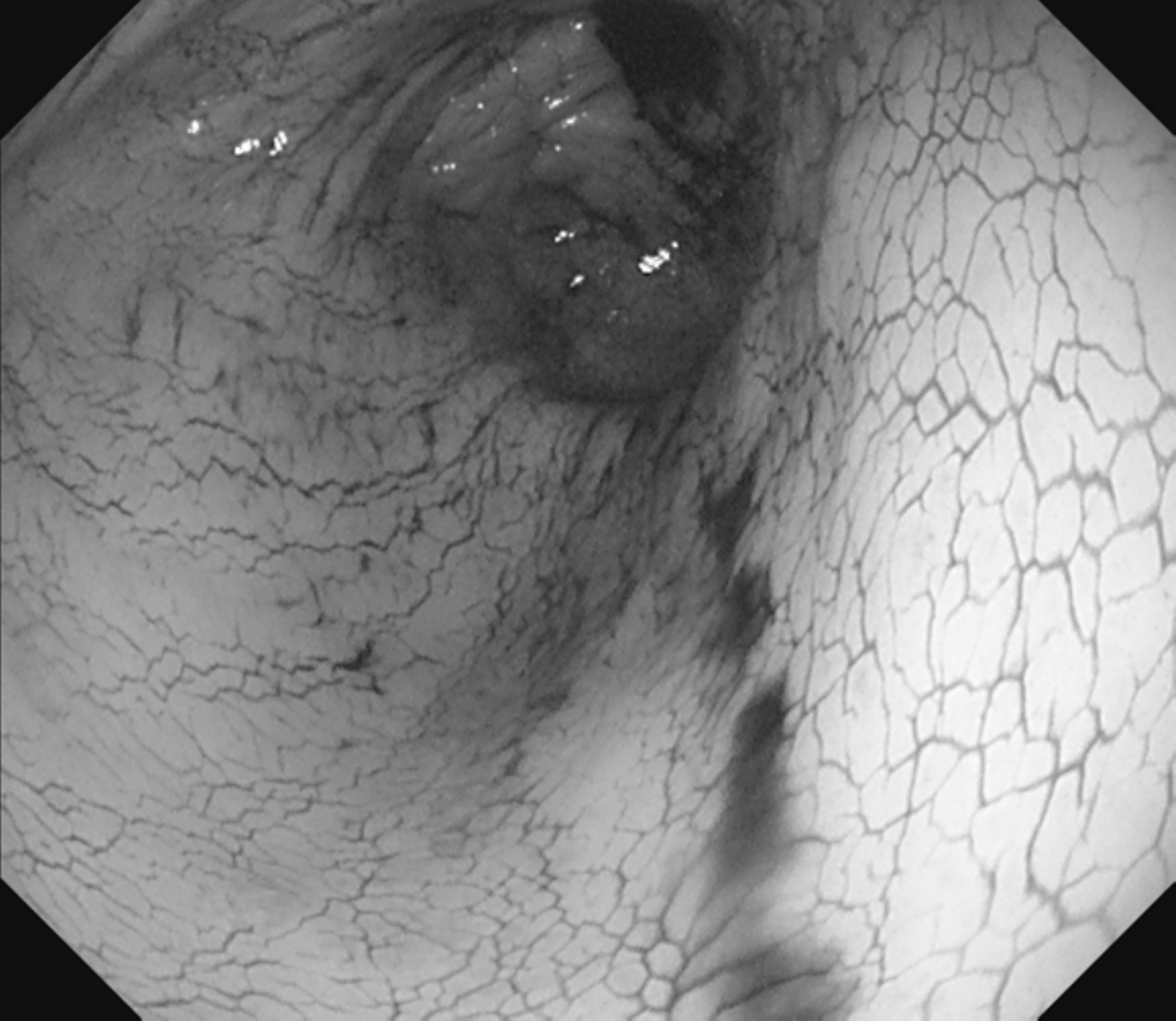}\\ \vspace{0.5mm}
    \caption{Red of (b)}
    \end{subfigure}\hspace{\fill}
    \begin{subfigure}{0.3\columnwidth}
    \centering
    \vspace{1mm}
    \includegraphics[width=\columnwidth]{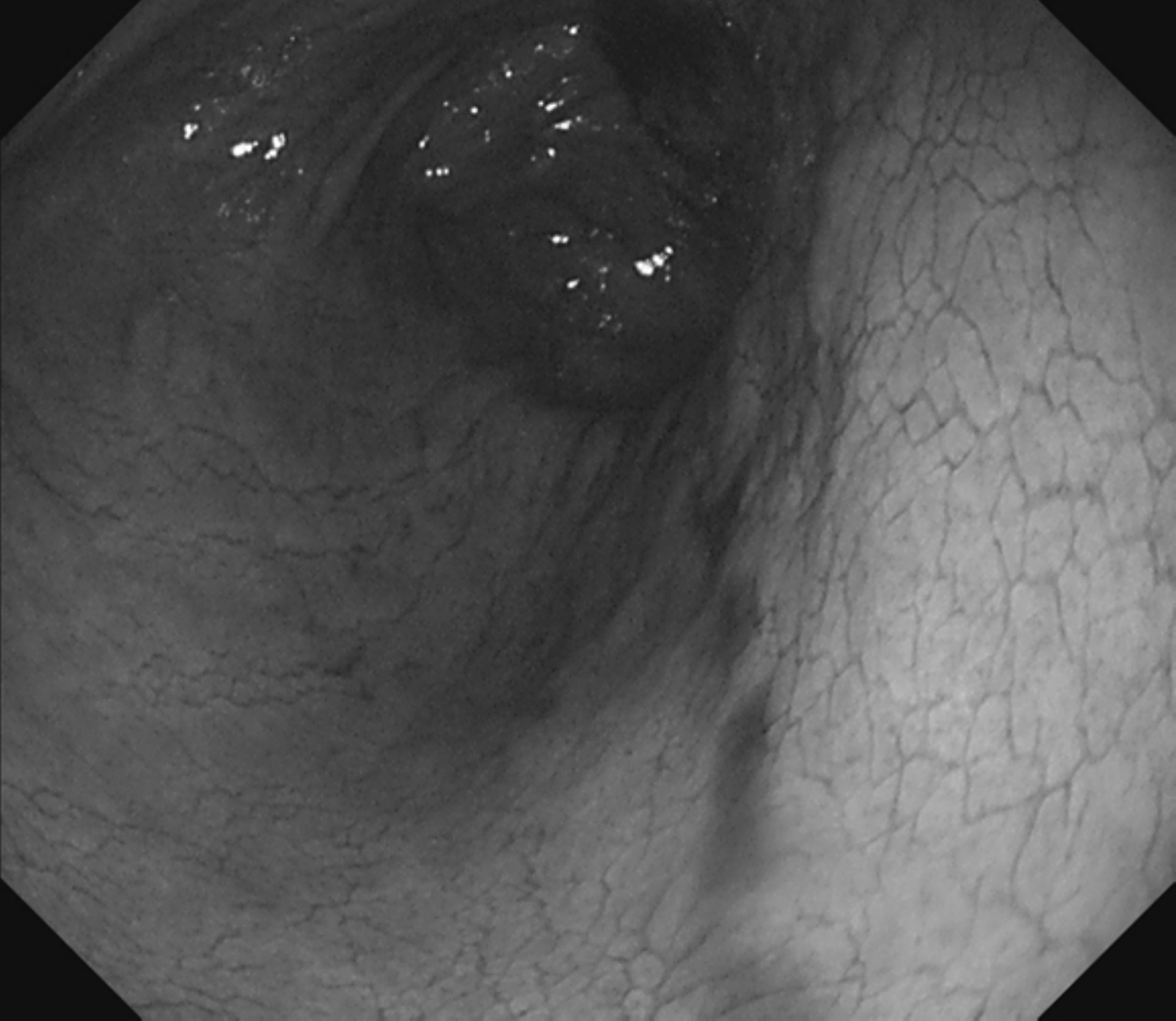}\\ \vspace{0.5mm}
    \caption{Green of (b)}
    \end{subfigure}\hspace{\fill}
    \begin{subfigure}{0.3\columnwidth}
    \centering
    \vspace{1mm}
    \includegraphics[width=\columnwidth]{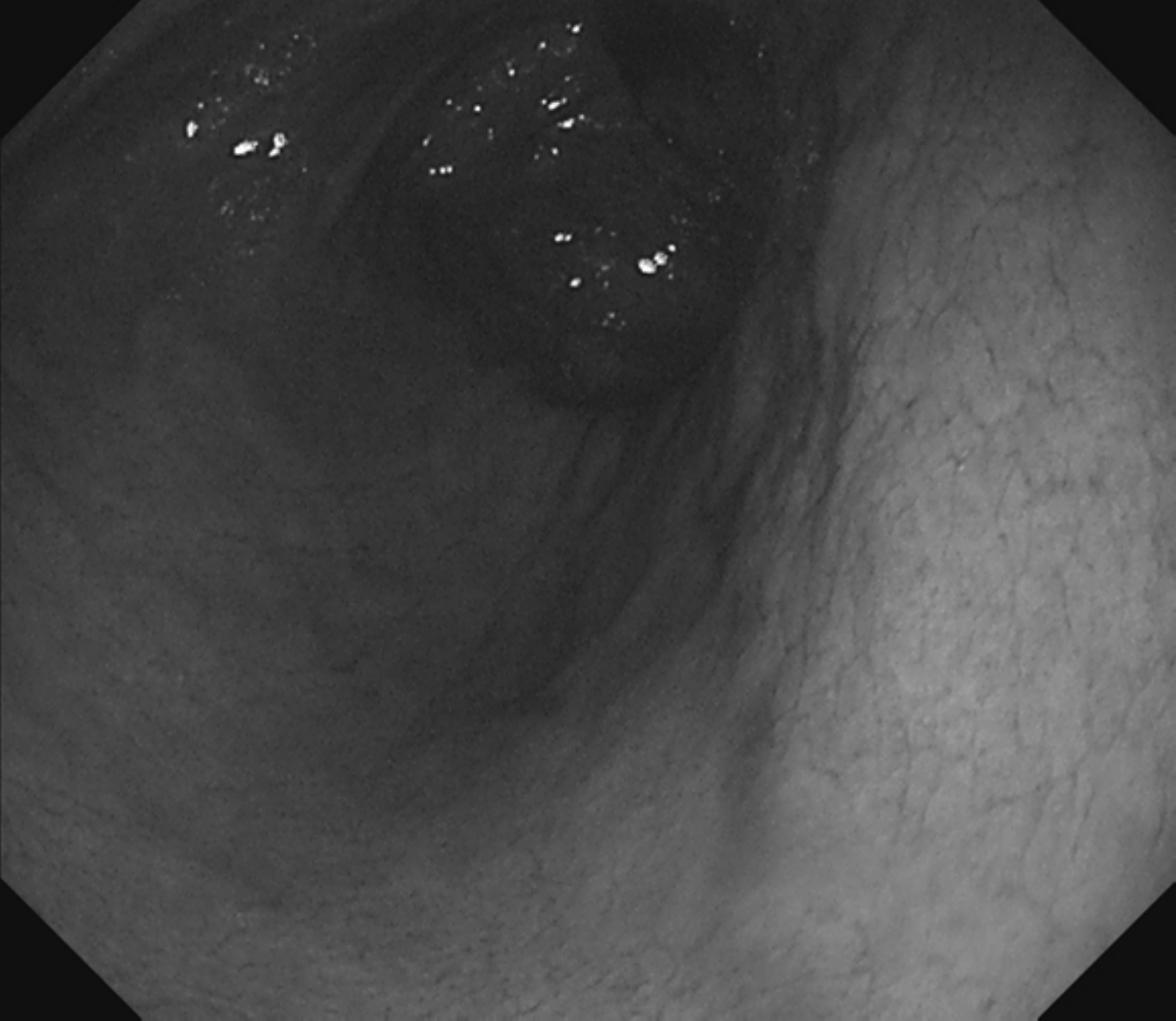}\\ \vspace{0.5mm}
    \caption{Blue of (b)}
    \end{subfigure}%
\caption{Examples of endoscope images captured without IC dye~(a) and with IC dye~(b). The color channel misalignment problem is observed in (a) and (b). The images (c)-(h) are six channel images extracted from (a) and (b). We can observe that the IC dye adds textures on the stomach surface, especially in the red channel~(f).}
\vspace{-1.2em}
\label{fig:colorartifact}
\end{figure}

\begin{figure*}[t!]
\centering
    \begin{subfigure}{0.3\textwidth}
    \centering
    \includegraphics[width=\columnwidth]{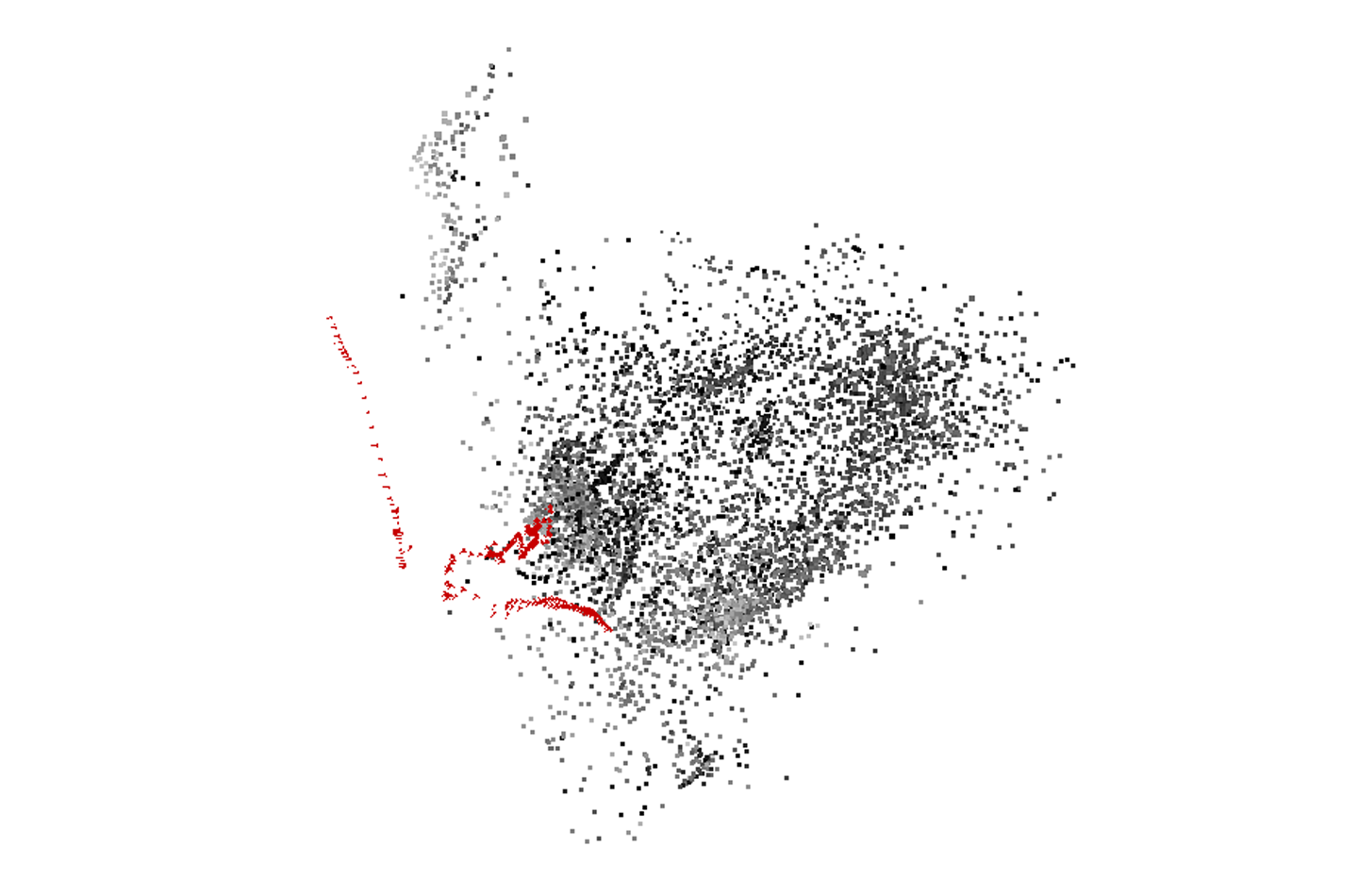}
    \caption{Red without IC dye}
    \end{subfigure}
    \begin{subfigure}{0.3\textwidth}
    \centering
    \includegraphics[width=\columnwidth]{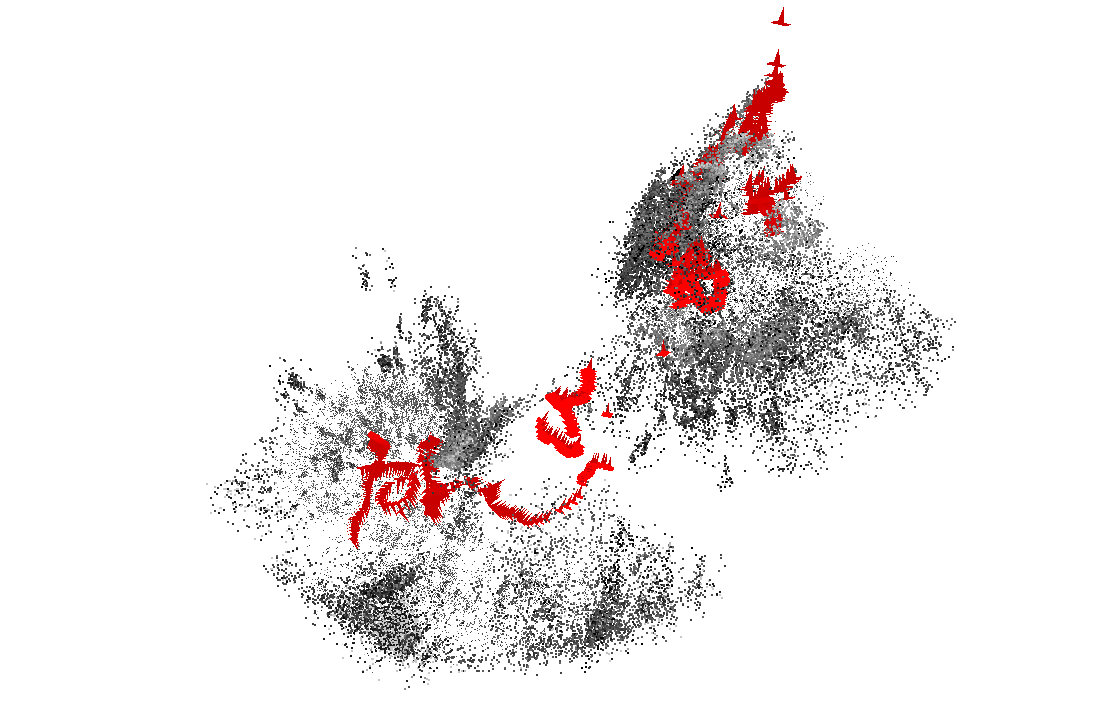}
    \caption{Green without IC dye}
    \end{subfigure}
    \captionsetup[subfigure]{aboveskip=6pt,belowskip=0.5pt}
    \begin{subfigure}{0.3\textwidth}
    \centering
    \includegraphics[width=\columnwidth]{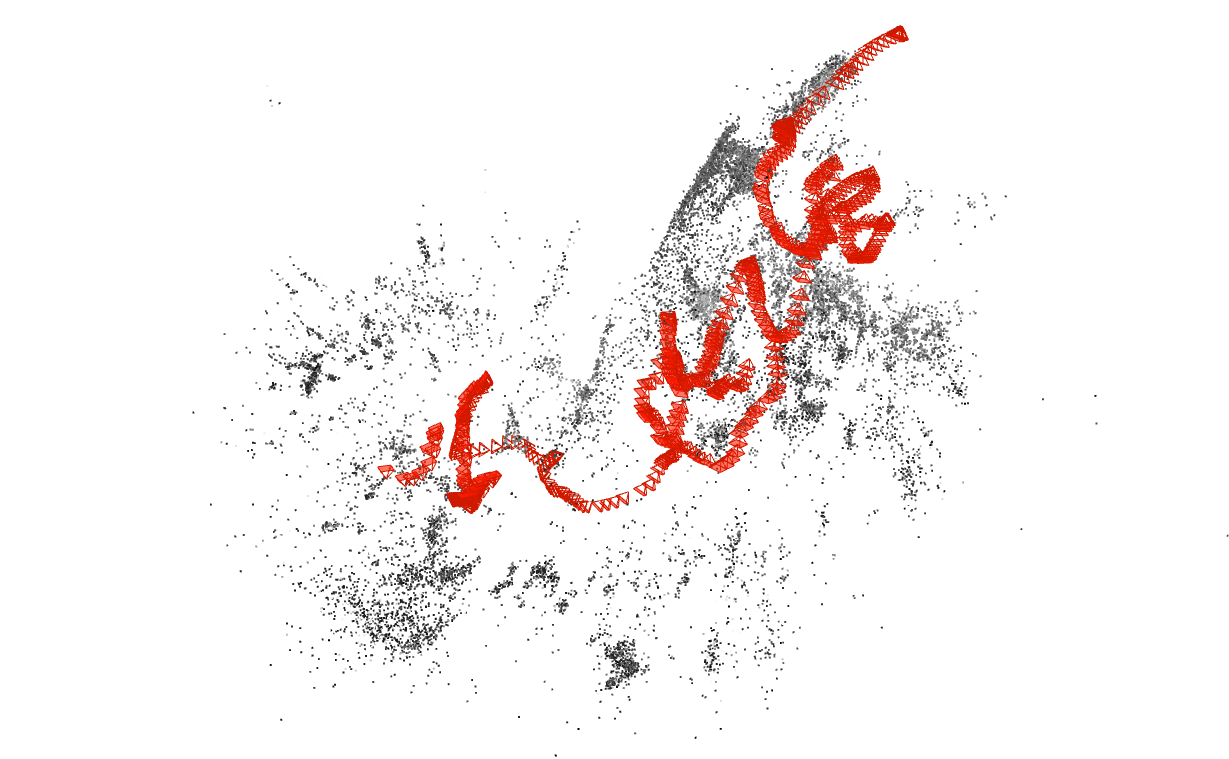}
    \caption{Blue without IC dye}
    \end{subfigure}
    
    \captionsetup[subfigure]{aboveskip=0.5pt,belowskip=0.5pt}
    \begin{subfigure}{0.3\textwidth}
    \centering
    \vspace{2mm}
    \includegraphics[width=\columnwidth]{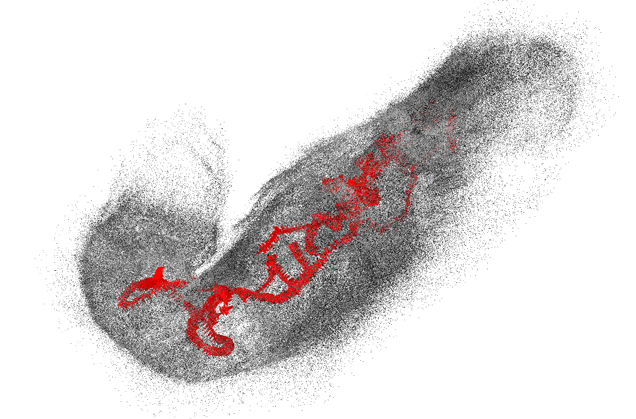}
    \caption{Red with IC dye}
    \end{subfigure}
    \begin{subfigure}{0.3\textwidth}
    \centering
    \vspace{2mm}
    \includegraphics[width=\columnwidth]{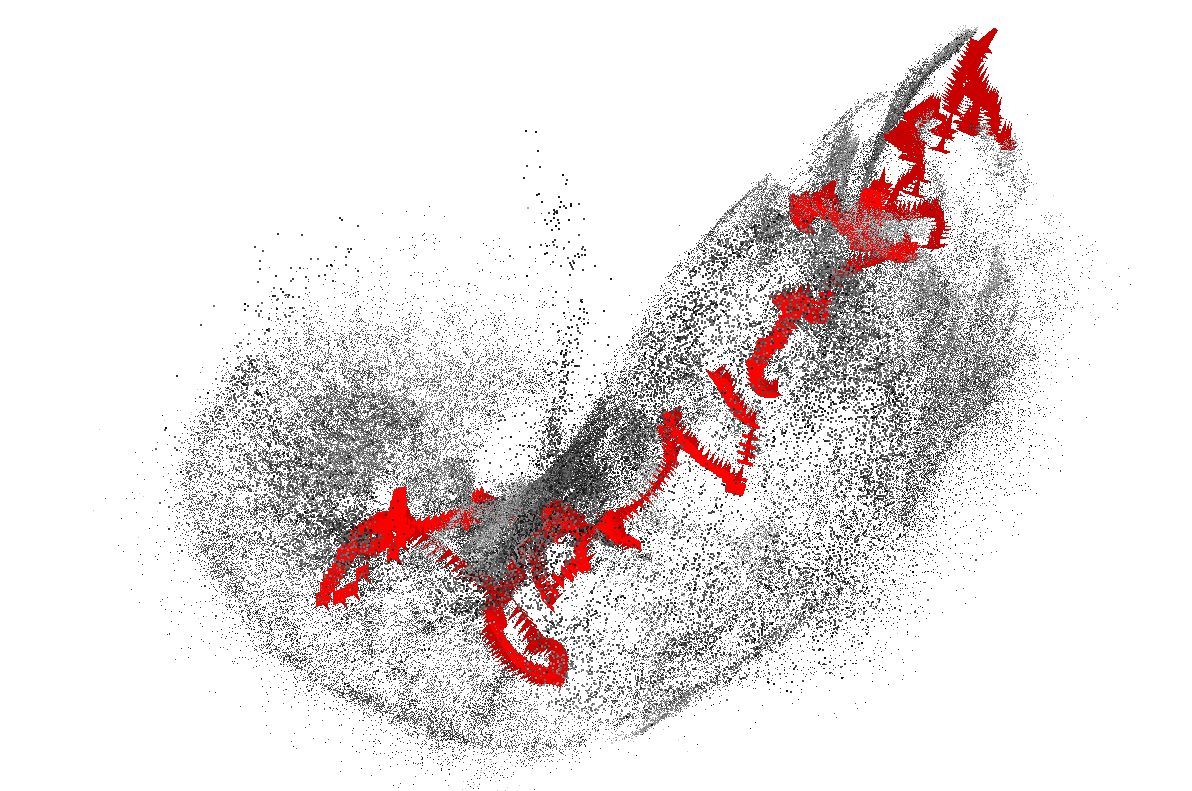}
    \caption{Green with IC dye}
    \end{subfigure}
    \begin{subfigure}{0.3\textwidth}
    \centering
    \vspace{2mm}
    \includegraphics[width=\columnwidth]{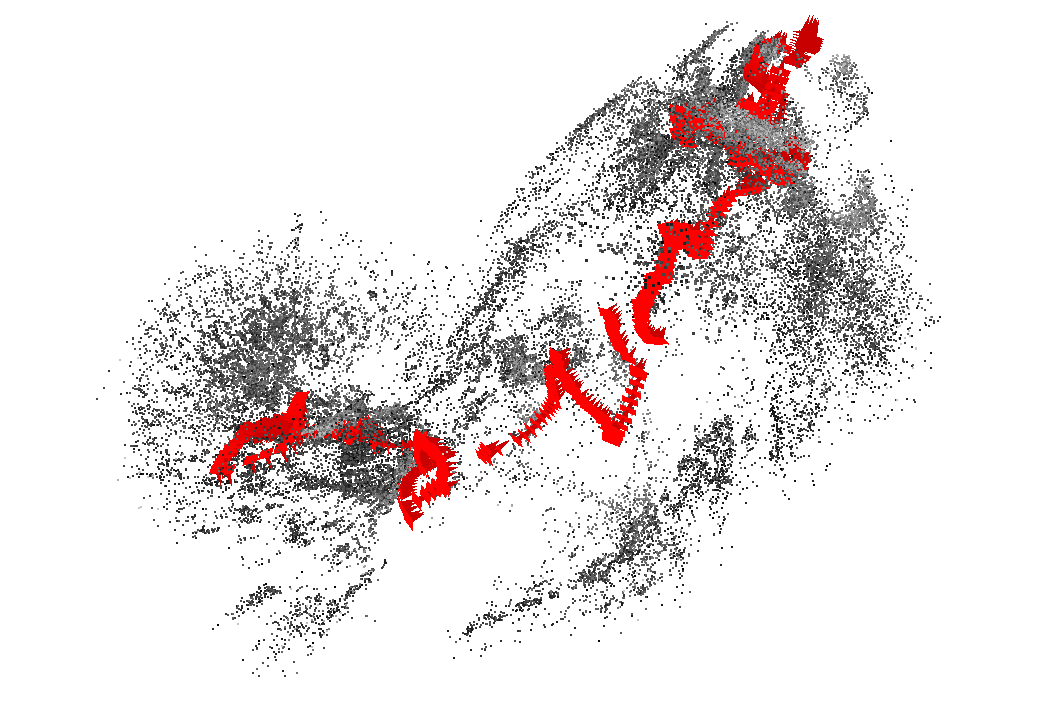}
    \caption{Blue with IC dye}
    \end{subfigure}
    
    \caption{The 3D point cloud results on Subject A. The gray dots represent the reconstructed 3D points and the red pyramids represent the estimated endoscope poses. There is a significant difference between the cases with and without IC dye. Only a sparse and small part of the stomach can be reconstructed in the case of without IC dye. On the other hand, the whole stomach can be reconstructed using the red channel with IC dye.}
    \label{fig:reconresult}
\end{figure*}

\begin{table*}[t!]
\centering
\caption{The objective evaluation of the 3D point cloud results using each color channel without and with IC dye.}
\label{tbl:extractedfeature}
\begin{adjustbox}{max width=\textwidth}
\renewcommand{\arraystretch}{2}
\begin{tabular}{c@{\hspace{2ex}}|c@{\hspace{2ex}}|c@{\hspace{2ex}}c@{\hspace{2ex}}c@{\hspace{2ex}}|c@{\hspace{2ex}}c@{\hspace{2ex}}c@{\hspace{2ex}}|c@{\hspace{2ex}}c@{\hspace{2ex}}c}
\hline
                                 &                     & \multicolumn{3}{c}{Subject A}                              & \multicolumn{3}{|c}{Subject B}                           & \multicolumn{3}{|c}{Subject C}                               \\
                                 &                     & Red                  & Green               & Blue          & Red                  & Green               & Blue       & Red                  & Green                & Blue          \\ \hline
\multirow{4}{*}{\shortstack{Without\\IC dye}} & Input images         & 2980                 & 2975                & 3000 & 734         & 729                 & 731        & 2959                 & 2790                 & 2977 \\
                                 & Reconstructed images & 207 (6.9\%)            & \textbf{687 (23.1\%)} & 497(16.6\%)    & 177(24.1\%)           & \textbf{226 (31.0\%)} & 138 (18.9\%) & 1064 (36.0\%)          & \textbf{1142 (40.9\%)} & 946 (31.8\%)    \\
                                 & 3D points            & 5740                 & \textbf{70610}      & 22764         & 8252                 & \textbf{15319}      & 3117       & 47960                & \textbf{79733}       & 40073         \\
                                 & Average observation & 173              & \textbf{670}     & 282       & 385              & \textbf{509}    & 158      & 329               & \textbf{467}     & 283       \\ \hline
\multirow{4}{*}{\shortstack{With\\IC dye}}  & Input images         & 1483                 & 1489       & 1476          & 2329                 & 2331       & 2319       & 2327        & 2304                 & 2323          \\
                                 & Reconstructed images & \textbf{1481 (99.8\%)} & 1249 (83.9\%)         & 567 (38.4\%)    & \textbf{2246 (96.4\%)} & 1488 (63.8\%)         & 335 (15.3\%) & \textbf{2297 (98.7\%)} & 891 (38.7\%)           & 361(15.5\%)    \\
                                 & 3D points            & \textbf{731070}      & 359418              & 49982         & \textbf{515762}      & 100114              & 12035      & \textbf{727954}      & 152223               & 14022         \\
                                 & Average observation & \textbf{4188}     & 1988              & 487       & \textbf{1971}     & 503              & 207    & \textbf{2656}     & 1195              & 221       \\ \hline

\end{tabular}
\end{adjustbox}
\end{table*}

In the input images extraction process, we first extracted all RGB frames from each video. Then, we extracted two image sequences from each video, where the first one consists of the images captured without IC dye (see Fig.~\ref{fig:colorartifact}(a)), while the second one consists of the images captured with IC dye (see Fig.~\ref{fig:colorartifact}(b)). After an in-depth inspection, we found that there are many color artifacts in the RGB images caused by color channel misalignment as shown in Fig.~\ref{fig:colorartifact}(a) and (b). To minimize the effect of the artifacts, we decided to use single channel images as SfM inputs. We also removed any duplicate frames that have almost no movement between successive frames. We used six channel images, as shown in Fig.~\ref{fig:colorartifact}(c)-(h), as SfM inputs and investigated the combined effect of chromo-endoscopy and color channel selection.

\subsection{Stomach 3D reconstruction}\label{sec:reconstruct}
The stomach 3D reconstruction follows the general flow of an SfM pipeline, assuming that the stomach has minimum movements. The algorithm starts with extracting features from the input images, matching the extracted features, and followed by the endoscope poses estimation and the feature points triangulation in parallel. This step generates a sparse point cloud of the stomach based on the endoscope motion and estimates each frame's pose with respect to each other.

We used SIFT~\cite{lowe_distinctive_2004} for feature extraction and exhaustively search to find the feature correspondences among all image pairs. We also applied bundle adjustment~\cite{triggs1999bundle} to optimize the 3D points and the endoscope poses.

\subsection{Mesh and texture generation}\label{sec:meshrecon}

Mesh and texture representation enables better visualization of the reconstructed 3D model. Our mesh generation starts by downsampling the original point cloud from the SfM result to a $n$ number of 3D points and removes outlier points using statistical outlier removal to generate a smooth mesh. The outlier removal starts by calculating the each 3D point's mean distance ($\bar{x_p}$) with its $n \times 0.1$ closest neighboring points. Assuming the distance distribution is Gaussian, the global distance mean ($\bar{x_g}$) and the standard deviation ($\sigma_g$) are then computed. Any 3D points whose mean distance $\bar{x_p}$ is over a threshold $\bar{x_g}+2\sigma$ are removed as outliers, leaving $k$ numbers of inlier 3D points. Then, the normal of each inlier 3D point is estimated based on its $k \times 0.1$ closest neighboring points. Each of the estimated normal is further refined using the related endoscope camera poses information to prevent it pointing outward. Finally, the triangle mesh is generated based on the outlier-removed 3D point cloud and its per-point estimated normal by using screened Poisson surface reconstruction~\cite{kazhdan2013screened}. To add more visual detail and functionality, we then applied a texture to the generated mesh based on the registered endoscope cameras in the SfM step. For each triangle mesh, we searched the best registered image for texturing based on the triangle-to-camera angle and distance.

\begin{figure*}
\centering
    \captionsetup[subfigure]{aboveskip=3pt,belowskip=0.5pt}
    \begin{subfigure}{0.49\textwidth}
    \centering
    \includegraphics[width=0.49\columnwidth]{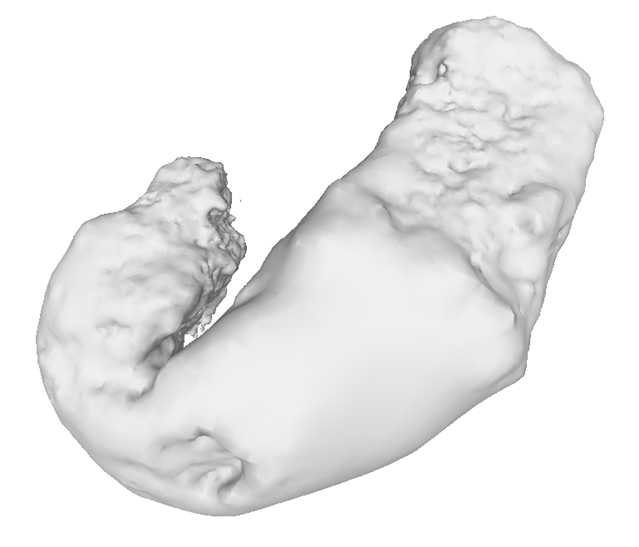}
    \includegraphics[width=0.49\columnwidth]{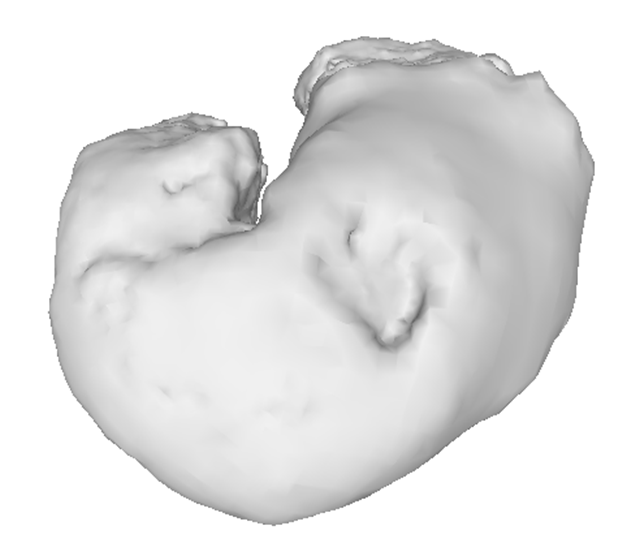}\\
    \vspace{2mm}
    \includegraphics[width=0.49\columnwidth]{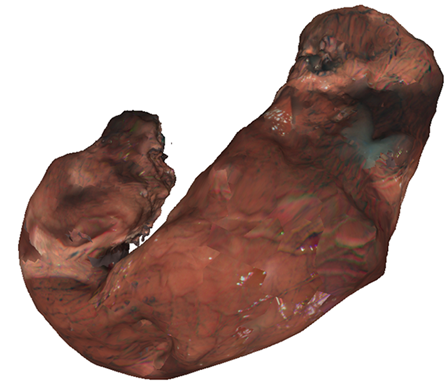}
    \includegraphics[width=0.49\columnwidth]{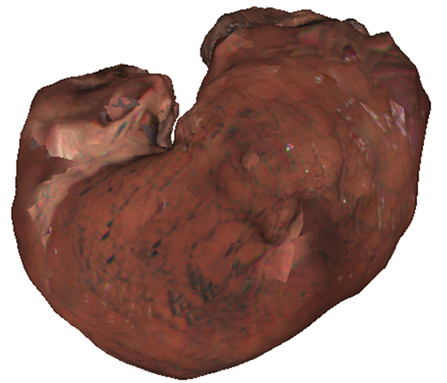}
    \begin{minipage}{0.49\columnwidth}
    \centering
    \small{View 1 \vspace{1mm}}
    \end{minipage}
    \begin{minipage}{0.49\columnwidth}
    \centering
    \small{View 2 \vspace{1mm}}
    \end{minipage}
    \caption{Subject B - Generated mesh and texture}
    \end{subfigure}
    \captionsetup[subfigure]{aboveskip=3pt,belowskip=0.5pt}
    \begin{subfigure}{0.49\textwidth}
    \centering
    \includegraphics[width=0.49\columnwidth]{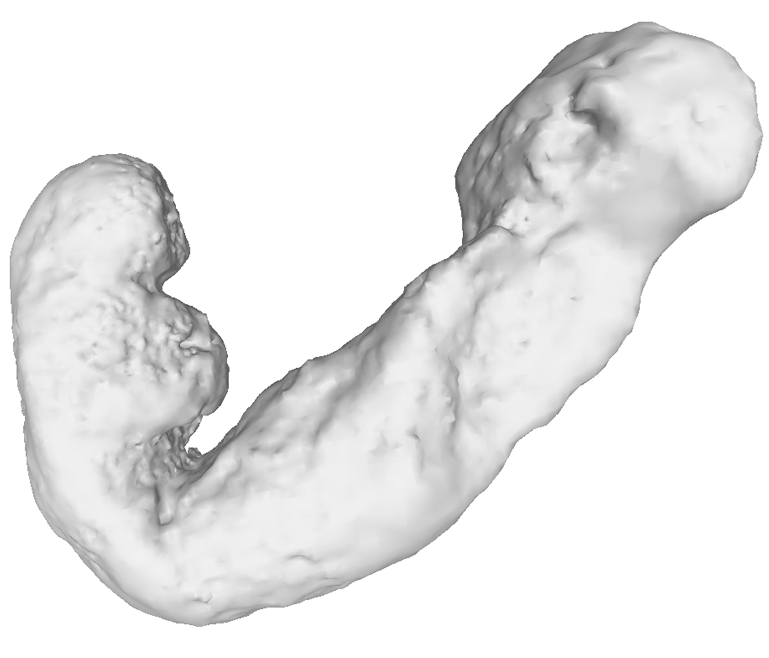}
    \includegraphics[width=0.49\columnwidth]{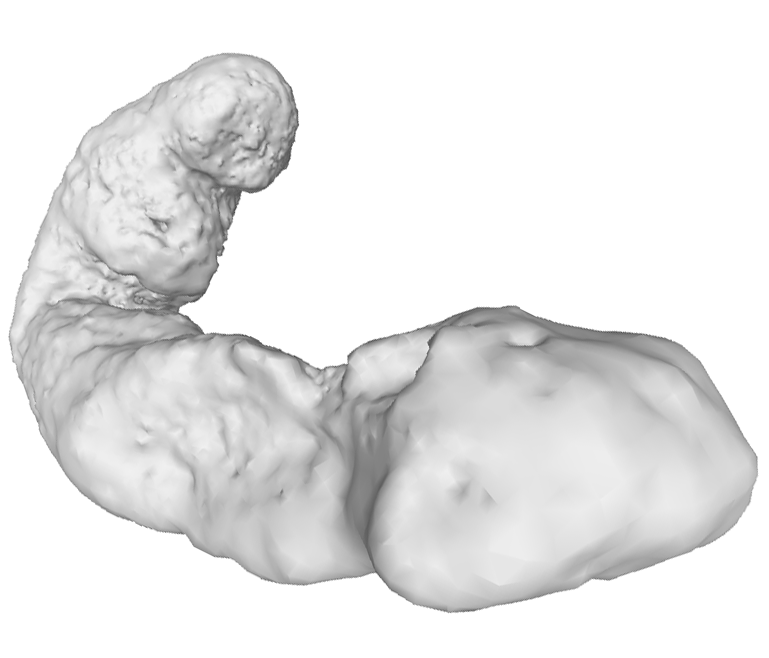}\\
    \vspace{2mm}
    \includegraphics[width=0.49\columnwidth]{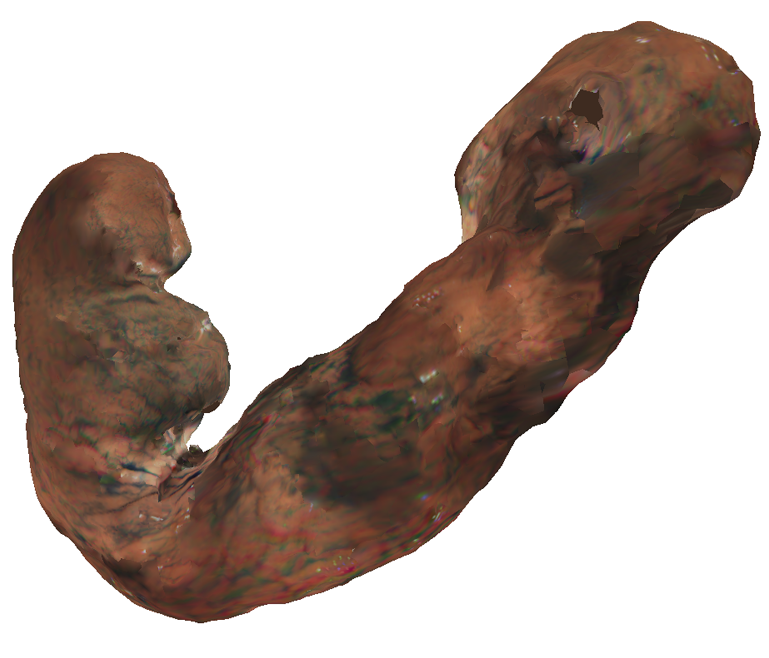}
    \includegraphics[width=0.49\columnwidth]{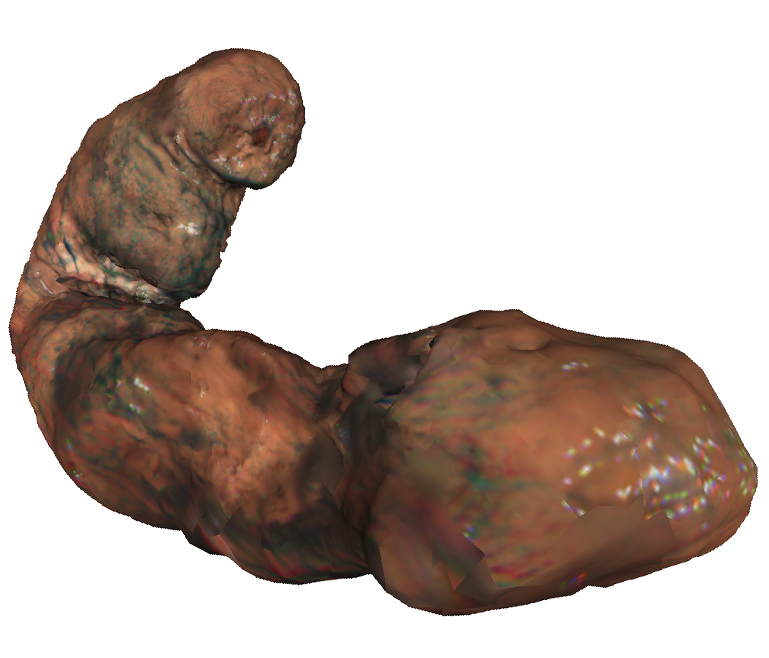}
    \begin{minipage}{0.49\columnwidth}
    \centering
    \small{View 1 \vspace{1mm}}
    \end{minipage}
    \begin{minipage}{0.49\columnwidth}
    \centering
    \small{View 2 \vspace{1mm}}
    \end{minipage}
    \caption{Subject C - Generated mesh and texture}
    \end{subfigure}
\caption{The triangle mesh and texture models generated from the point clouds reconstructed using the red channel with IC dye. The shown texture is the inner texture of the stomach. The video version can be seen from the following link (\textcolor{blue}{http://www.ok.sc.e.titech.ac.jp/res/Stomach3D/}).}
\label{fig:meshreconresult}
\end{figure*}

\begin{figure*}
    \begin{subfigure}{0.49\textwidth}
    \centering
    \vspace{3mm}
    \includegraphics[width=\columnwidth]{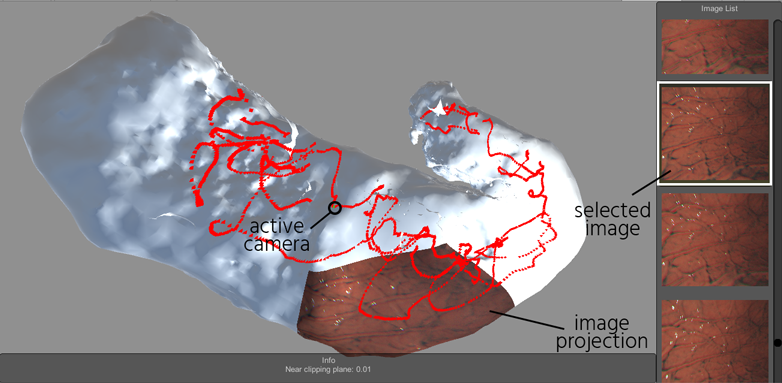}\\ \vspace{1mm}
    \caption{Our custom viewer interface}
    \end{subfigure}
    \hfill
    \begin{subfigure}{0.49\textwidth}
    \centering
    \vspace{3mm}
    \includegraphics[width=\columnwidth]{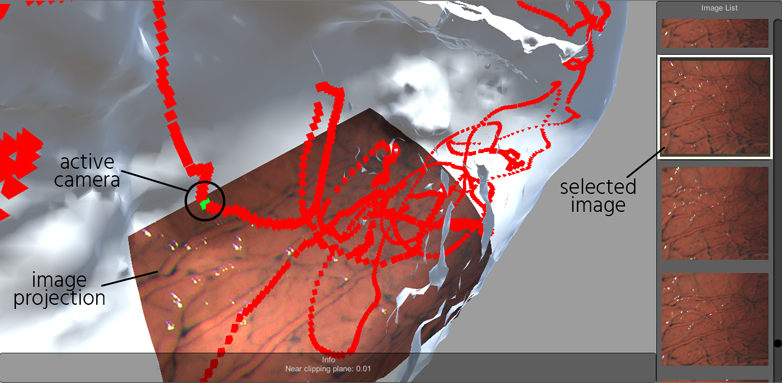}\\ \vspace{1mm}
    \caption{Image projection of the selected image to the mesh}
    \end{subfigure}
    \caption{Demonstration of our custom viewer. The figure (a) illustrates the viewer interface. It loads the generated mesh and the estimated endoscope camera positions. The red pyramids in (a) and (b) represent the estimated cameras and the camera trajectory. The user can select either a camera or an image to project the related image to the mesh, as shown in (b). The selected or active camera is shown as green in both (a) and (b).}
    \label{fig:viewer}
\end{figure*}

\section{Results and Discussion}

We performed the endoscope camera calibration using the OpenCV camera calibration library. The SfM pipeline was implemented on Colmap~\cite{schonberger_structure--motion_2016}. For filtering the point cloud, we set as $n = 10000$ to generate a smooth triangle mesh. We applied screened Poisson reconstruction~\cite{kazhdan2013screened} for triangle mesh generation. For the texturing purpose, we applied the parameterization and texturing function from Meshlab~\cite{MeshLab51:online}.

Figure~\ref{fig:reconresult} shows the 3D point cloud results on subject~A, which are reconstructed using different color channels of the cases without and with IC dye. In general, the channels with the IC dye (Fig.~\ref{fig:reconresult}(d)-(f)) give a more complete reconstruction result compared to the channels without the IC dye (Fig.~\ref{fig:reconresult}(a)-(c)). In the case without the IC dye, the green channel gives the best result even though the model is full of holes. In the case with the IC dye, the results of all the three channels have the shape of the stomach. Among the RGB channels, the red channel gives the most complete and densest result. Some holes still exist in the result using the green channel, while the blue channel is only able to reconstruct around $3/4$ of the whole stomach.

Table~\ref{tbl:extractedfeature} shows the objective evaluation of the 3D point cloud results on all three subjects. Table~\ref{tbl:extractedfeature} shows that the number of 3D points is generally higher when the IC dye is present. We also notice that the average observation, which represents the per-image average number of the 2D feature points that can be triangulated into the 3D points, is generally increased when the IC dye exists. In addition, the percentage of reconstructed images over input images is significantly increased by using the IC dye. Among all the results, the red channel with the IC dye gives the best result, where more than 95\% images are reconstructed. When the IC dye is not present, the green channels gives the best result.

The above subjective and objective evaluation consistently shows that the red channel with the IC dye gives the best result. As shown in Fig.~\ref{fig:colorartifact}(c)-(h), this is because that the red channel leverages the effect of the IC dye more than the other channels. In~Fig.\ref{fig:colorartifact}(f), many textures, from which many distinctive features can be extracted, are apparent in the red channel. When the IC dye is not used, the green channel has better contrasts compared to the others channels. The blue channel is the least preferable among those three channels for both cases without and with the IC dye.

Figure~\ref{fig:meshreconresult} shows the results of triangle mesh and texture generation using the red channel with the IC dye. We can confirm that the generated meshes resemble the whole shape of a stomach. Moreover, the textured representation makes the generated 3D model more perceptible for viewers.

Figure~\ref{fig:viewer} shows our custom viewer that can project any selected reconstructed images to the generated triangle mesh based on the estimated endoscope poses in SfM. This custom viewer provides viewers with the estimated location of a particular image frame, which can be used for the 3D localization of a malignant lesion. Our viewer should be very valuable for gastric surgeons to make a medical decision. 

\section{Conclusion}

In this paper, we have presented an offline solution to reconstruct the whole shape of a stomach from a standard monocular endoscope video. To obtain better reconstruction quality using SfM, we used a single channel images without color channel misalignment artifact. We found that the chromo-endoscopy with IC blue color dye generally gives significant improvement to the completeness of the reconstruction result. Furthermore, we found that the red channel with the IC dye provides the most complete 3D model compared to the other channels. A custom viewer that can localize a particular image frame in the reconstructed 3D model was also presented. In future work, we plan to refine the mesh generation process for more detail representation considering more effective downsampling and outlier removal approaches. To view the results in more detail, please visit our project page in the following link (\textcolor{blue}{http://www.ok.sc.e.titech.ac.jp/res/Stomach3D/}).

\bibliographystyle{IEEEtran}

\end{document}